\documentclass[conference]{IEEEtran}

\usepackage{graphicx}
\usepackage{hyperref}
\usepackage[sort&compress,numbers]{natbib} 
\graphicspath{ {./images/} }

\usepackage{tabularx,booktabs}
\newcolumntype{C}{>{\centering\arraybackslash}X} 
\title{Who's a Good Boy? Reinforcing Canine Behavior in Real-Time using Machine Learning}

\author{
\IEEEauthorblockN{Jason Stock}
\IEEEauthorblockA{Dept. Computer Science\\
Colorado State University\\
\href{mailto:stock@colostate.edu}{stock@colostate.edu}}
\and
\IEEEauthorblockN{Tom Cavey}
\IEEEauthorblockA{Dept. Computer Science\\
Colorado State University\\
\href{mailto:tomcavey@colostate.edu}{tomcavey@colostate.edu}}
}

\begin{document}
\maketitle
\thispagestyle{plain}
\pagestyle{plain}

\begin{abstract}
In this paper we outline the development methodology for an automatic dog treat dispenser which combines machine learning and embedded hardware to identify and reward dog behaviors in real-time. Using machine learning techniques for training an image classification model we identify three behaviors of our canine companions: ``sit'', ``stand'', and ``lie down'' with up to 92\% test accuracy and 39 frames per second. We evaluate a variety of neural network architectures, interpretability methods, model quantization and optimization techniques to develop a model specifically for an NVIDIA Jetson Nano. We detect the aforementioned behaviors in real-time and reinforce positive actions by making inference on the Jetson Nano and transmitting a signal to a servo motor to release rewards from a treat delivery apparatus.
\end{abstract}

\section{Introduction}
Dog trainers have mastered the skill of teaching obedience by rewarding desired actions with food or auditory queues. However, as identified in \cite{Gerencsr2018DevelopmentAV, Chiandetti2016CanCT}, the learned behavior of dogs may diminish in early stages of training when actions go unrewarded. To maximize the efficacy of learning, we model the actions of a dog trainer with machine learning to identify behaviors and reinforce commands such as ``sit'' or ``lie down'' in real-time. We achieve this by utilizing a neural network on a NVIDIA Jetson Nano equipped with a camera and a mechanism for delivering treats. 

Through extensive experiments we evaluate how different neural network architectures designed for image classification perform on an embedded device. Moreover, we compare and contrast performance of post training quantization methods available in TensorFlow's model optimization toolkit with that of TensorRT: NVIDIA's highly optimized C++ library for deep learning inference. By optimizing the model efficiently we can reduce the memory footprint, computational power usage, CPU and hardware accelerator latency with minimal loss in model performance. We find that neural networks, when developed correctly, perform significantly better but carry notable compromises. Specifically, when using TensorFlow to quantize weights it may be better to incur a marginal increase of memory utilization with a corresponding increase in accuracy or at the expense of a greater model size. 

Ultimately, we discern two primary use cases and results for our applications, namely: (i) using TensorRT to optimize the EfficientNetB0 model to 16-bit weights for inference at 36.28 frames per second (FPS) with 90.75\% test accuracy and a model size of 28.34 MB, and (ii) using post training 16-bit floating point quantization from TensorFlow on MobileNetV2 for inference at 9.38 FPS with 87.57\% test accuracy and a binary model size of 4.48 MB. We individually deploy and evaluate each model on the apparatus with a Australian Shepherd as our test subject. Within seconds, the Jetson Nano is making inference and the subject begins to show interest and is appropriately rewarded. While application (i) is exceptional on paper, we determine that the use of model optimizations with TensorFlow, and use case (ii), to be more stable and consistent.

\subsection{Related Works}

The behavior and identification of dogs have been studied in various domains that include: veterinarian science \cite{Weiss2015CanineBL, Mundell2020AnAB}, human-computer interaction \cite{Majikes2017BalancingNS, Brugarolas2013BehaviorRB}, and computer vision \cite{Raduly2018DogBI, Ehsani2018WhoLT, Zamansky2019AnalysisOD, Cao2019CrossDomainAF, Randhavane2019LearningPE}. We find that many existing works focus on capturing the behavior of dogs with the inclusion of wearable sensors. In \cite{Majikes2017BalancingNS}, the posture of a dog, e.g. sitting, standing, or eating, as captured by an accelerometer on a wearable instrument is classified via random forests and an algorithm which uses a variance-based threshold to separate classes. Similarly, Brugarolas \textit{et al.} \cite{Brugarolas2013BehaviorRB} use decision trees and hidden markov models to detect static and dynamic behaviors from worn accelerometers and gyroscopes. It is evident that utilizing wearable device has shown success with the use of machine learning. Although, we extend these works by using a different mode of environmental recognition, namely image classification via convolutional neural networks. 

Tasks that utilize computer vision exist for works similar to dog breed classification and analyzing sleeping behaviors, however; they are often developed for systems with large resource availability. For example, Raduly \textit{et al.} \cite{Raduly2018DogBI} use the camera on a smart phone to capture images to classify dog breeds. The NASNet-A network architecture is used for on device inference, but achieves a low accuracy relative to a more complex network. Therefore, to reach higher accuracies, a remote server is used for online queries to the ResNetV2 model. The drawback to this approach is that processing time is constrained to network communication and requires an active network connection. As a result, we provide insights to how a model can be developed for offline inference for computer vision applications with near real-time results.

In general, this work separates the need for specialized hardware by leveraging computer vision techniques to make inference about a scene with a camera alone. The ability to identify the behavior using only computer vision is a challenging task. However, Weiss \textit{et al.} \cite{Weiss2015CanineBL} illustrate the unique characteristics in dogs that are visually noticeable. For example, the subtle difference between offensive and defensive aggression can be observed by the position of the tail or angle of the ears. Specialized machine learning models benefit from these visual cues in that identifying edges and unique behaviors and patterns, thus, further motivating the use of neural networks in this study.

Perhaps the most similar work comes from Moundell \textit{et al.} \cite{Mundell2020AnAB} who incorporates a microphone and camera to identify posture as well as positions, trajectory and vocalizations of dogs. However, in the present study we largely focus on how to best design a neural network for our application on an embedded device, which is not discussed in the related work. Developing a model to fit on low cost systems introduces an additional layer of complexity that we assess through extensive experimentation and methodical decision making.

Many applications in machine learning and embedded systems are explored outside the area of artificial and behavioral dog training. A study from Muhammad \textit{et al.} \cite{Muhammad2019}, for example, leverage a Jetson Nano for real-time fire detection using a convolutional neural network with video captured on surveillance cameras. The authors utilize the MobileNetV2 \cite{Sandler2018MobileNetV2IR} architecture and dominate state-of-the-art methods, thereby motivating our use of the architecture. We find the Jetson Nano in use with other machine learning applications such as a real-time lane detection for autonomous driving \cite{Wang2019} and a mobile sign language translation \cite{Zhou2020}. These studies explore a range of network architectures and sufficiently demonstrate the use of the Jetson Nano for computer vision tasks. Therefore, we justify the device as an appropriate system to use in the present study. 

\subsection{Contributions}
Our contributions exemplify methods to use computer vision in aiding the identification of behavior in dogs on an embedded device. We provide insights into appropriate models suggested for use on low memory and low power devices such as the Jetson Nano. We provide direction on how to design and build the apparatus used for delivery of rewards. The project in its entirety is made available on GitHub and can be found at \url{https://github.com/stockeh/canine-embedded-ml}

\begin{figure}[t]
\includegraphics[width=8.5cm]{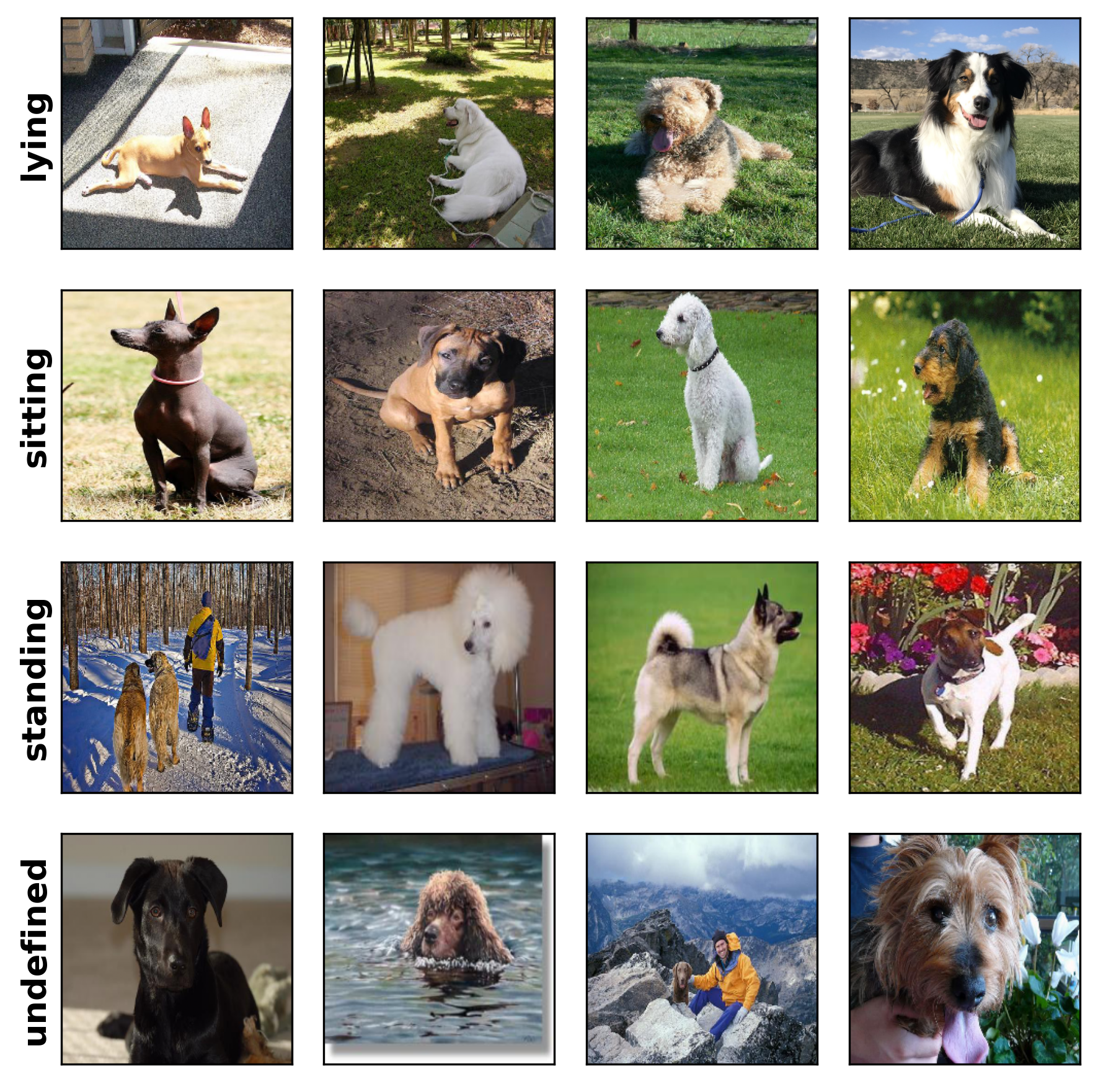}
\caption{Dataset samples corresponding to distinct behaviors identified in dogs.}
\centering
\label{fig:dataset_examples}
\end{figure}

\section{Dataset Details}
To the best of our knowledge, no dataset exists with images of dogs and corresponding labels pertaining to their actions. Therefore, we utilize the Stanford Dog Dataset \cite{KhoslaYaoJayadevaprakashFeiFei_FGVC2011} and hand-label a set of four actions: standing, sitting, lying down, and undefined, yielding 4143, 3038, 7090, and 6307 samples, respectively. \autoref{fig:dataset_examples} displays random samples drawn from the dataset that correspond to each behavior. As identified in the bottom row, indistinguishable actions, such as a close frame of a dog’s face, are labeled as undefined as there is no way to determine the behavior. The majority of samples that fall in this category are close portraits of the dog's face. This is not ideal for our application as even to the human eye it can be difficult to identify the behavior.

The 20,578 total RGB images contain 120 different dog breeds in different scenes with varying lighting conditions and resolutions. Not every image shares the same shape, but 50\% of the data lie between a resolution of 361x333 and 500x453. This allows for downsampled images to have a consistent dimension without losing much information or interpolating missing values. We further preprocess the data by means of augmentation in the training pipelines. Adding minimal changes to rotation and randomly reflecting an image about the horizontal is primarily done to artificially increase the size of the dataset and improve training. Together, the variety of samples is important to have a network that generalizes well to unforeseen environments and diverse dog breeds.

\section{Experimental Methodology}
\subsection{Models and Training Details}

We propose using a Convolutional Neural Network for image classification. A network architecture and hyperparameter search is performed over hand crafted models those inspired by state-of-the-art methods for mobile devices. Exploring the design space of hand crafted networks, named ConvNet, allow for starting with simple networks which include few parameters and incrementally increase the complexity. Thereafter, we evaluate MobileNetV2 \cite{Sandler2018MobileNetV2IR} and five variants of the EfficentNet (B0-4) \cite{Tan2019EfficientNetRM} architecture for use with fine-tuning and transfer learning from ImageNet. Additionally, we explore variations in data preprocessing and augmentation as well as investigate how image resolution influences performance. 

The machine we use only for training includes two Intel Xeon Silver 4216 CPU’s at 2.10GHz, two 24GB NVIDIA Quadro RTX 6000 GPU’s, and 256GB DDR4 memory. On average we find training ConvNet takes 14 seconds per epoch, MobileNetV2 takes 39 seconds per epoch, and EfficientNet takes between 54-101 seconds per epoch depending on the model variant being used. 

\subsection{Hardware and Apparatus Design}
The hardware we use to solve this problem includes the NVIDIA Jetson Nano and an array of external peripherals. The Jetson Nano is equipped with 128 NVIDIA Maxwell GPU cores, 4 GB of RAM, and powered with a 10 W switchable power supply for maximum performance. As supported by previous works, we find this to be more than enough computational power to run our hand designed model and our fine tuned transfer learning models. The Nano has analog and digital GPIO for use with external devices such as a camera, a servo motor, and a speaker. To control the servo motor we use a 16-bit PWM I2C driver (PCA9685) with a dedicated 5 W power supply. The Nano, I2C board, and the servo motor have small dimensions such that we can design an enclosure which can be easily portable.

\begin{figure}[t]
\includegraphics[width=8cm]{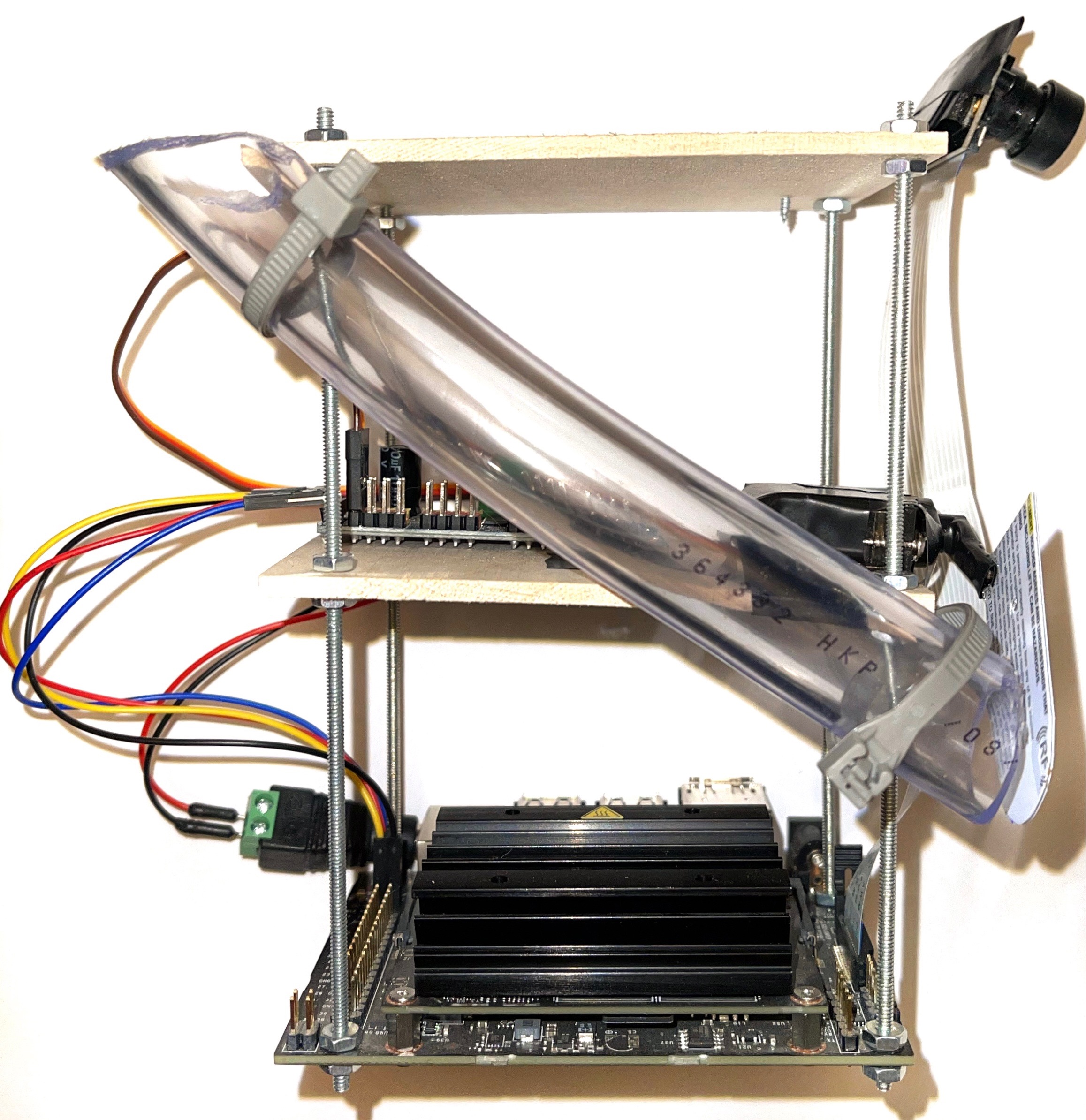}
\caption{NVIDIA Jetson Nano, PCA9685 PWM I2C driver, 180 degree servo motor, 8MP IMX219 180 degree camera, and dog treat delivery tube}
\centering
\label{fig:hardware_components}
\end{figure}

The apparatus is a prototype design made of balsa wood platforms and aluminum threaded rods. It takes a rectangular shape of roughly 8 inches height x 5 inches width x 5 inches length. At the top of the device, there is a forward facing camera, mounted on a hinge, which sits at a slightly downwards angle to ensure capturing activity of an animal on the ground. The inside of the enclosure houses the Jetson Nano and I2C board. The side of the apparatus contains a vertically mounted plastic tube, which is used as a container and delivery system for treats. At the bottom of the tube there is a plastic door attached to the servo motor arm. When the servo motor mechanism is activated the arm rotates 60 degrees to actuate the door up and down to release a treat. \autoref{fig:hardware_components} displays the apparatus assembled.  

\subsection{Post Training Quantization}
After evaluating network metrics for our best performing models, we focus on model compression and accelerating inference on the Jetson Nano by leveraging TensorRT optimizations and TensorFlow quantization. With TensorFlow we explore three primary means of post-training optimizations, namely: standard (lite) 32-bit floating point, 16-bit floating point, and dynamic range quantization. With TensorRT we consider optimizations of the TensorFlow saved model to 16/32-bit floating point precision. The methods are evaluated for each model and results are contrasted in detail.

The first two methods from TensorFlow are ideal for running on the GPU where the last is documented to run efficiently on the CPU. With the Nano's dedicated GPU, we find the first two methods to be advantageous as converting the weights to 16-bit floating point numbers complements the precision of the Jetson Nano's GPU. Although, we consider dynamic range in our experiments as it has room for the greatest decrease in memory footprint. This is due to weights dynamically being converted to 8-bits of precision based on their range, all the while keeping output values as floating point numbers; which provides latencies close to full fixed-point computations with the supported hardware. 

\section{Architecture Search and Hyperparameter Optimization}
Each model explored herein undergo an extensive architecture and hyperparameter search. The focus is to explore the design space to find a model that best represents the data and performs accurately and consistently on the Jetson Nano. An ideal model will have a high measure of accuracy, generalize well to images that have never been seen before, and perform efficiently on a low power accelerator. 

A naive approach may be to use a highly complex and deep neural network as typically, the more parameters a model has the more mapping functions it can represent. Thus, having more flexibility to learn progressively higher level features between the input to each output class. However, this also increases the variance of the model and the risk of overfitting to unrepresentative or noisy training data. Additionally, a larger model will incur a higher computational cost which is not optimal with use on an embedded device.

Coincidentally, using a simpler model (in terms of the number of parameters) may not capture the important features in the data or represent the desired mapping. This will result in a model with higher bias and training data could be underfit. Although, having a shallow network and/or fewer parameters will result in fewer computations and more efficient training and inference speeds. Considering these constraints, we would like to find a sufficient mapping that is complex enough to identify patterns in the dog's actions, but simple enough to run efficiently.

For all experiments, we partition the dataset into a training, validation, and testing set with 75\%, 10\% and 15\% of the data, respectively. Furthermore, we ensure that each dataset has an equal class distribution and we shuffle the samples to randomize the order of breeds seen during training. The dataset has a class imbalance with nearly 2x samples for standing over sitting images. This leads to the network prioritizing samples that are seen more often. To remedy this, we introduce data augmentation with random rotations of ±8 degrees, horizontal flips, and minor translations in width and height. As a result, we observe more stable results at the expense of slightly longer training times.

\subsection{Custom Implementation}
Finding a network structure for ConvNet is an iterative process with exploration over the layer topology, number of epochs, batch size, activation functions, optimization algorithms, and learning rates. Over each network search we preprocess the images to three different resolutions, including 32x32, 64x64, and 256x256. Using a lower dimension removes potentially redundant and duplicate values, and subsequently speeds up training time. However, with higher resolution we are able to assess deeper networks when using max pooling, and found better results with accuracy increasing by 5\% when using a resolution of 256x256 over 32x32.   

We run a breadth of experiments and find the best model to consist of four convolutional layers with 32, 64, 128, and 256 filters, each with a 3x3 kernel size and stride of one, from the input forward. Each convolutional layer is followed by batch normalization, ReLU activation function, max pooling with a pool size of two, and dropout with a probability of 0.20. Following is a single dense layer of 128 units with the ReLU activation, and a linear output layer of three units that then pass through the softmax activation to convert real numbers into a probability distribution consisting of three probabilities for each class (i.e., lying, sitting, and standing). The network optimizes the sparse categorical cross entropy using adaptive moment estimation (Adam) with a learning rate of 0.0001 over 35 epochs using a batch size of 32.   

\begin{figure}[t]
\includegraphics[width=8.5cm]{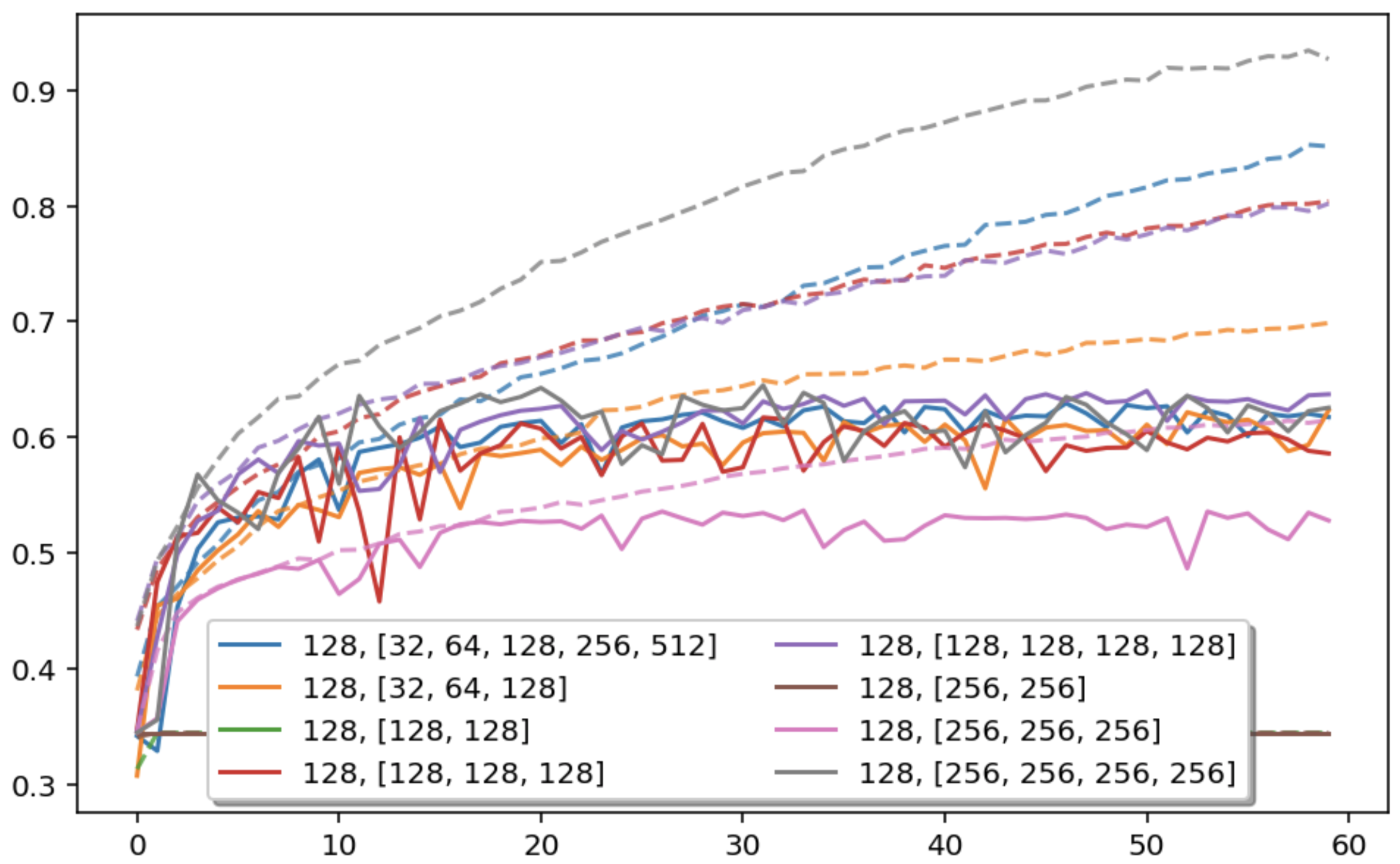}
\caption{Model accuracy over training epochs for the training set (dashed line) and validation set (solid line). All custom implementation models are trained using the same hyperparameters from a previous test for 60 epochs. The legend represents the dense units followed by a list of values with each number representing the number of filters in the layer from input to the dense layer.}
\centering
\label{fig:model_search_epochs}
\end{figure}

ConvNet achieves 69.95\% accuracy on the test set with lying and standing having the highest accuracy of correctly predicted samples. By observing a confusion matrix for each class we find that images labeled as sitting are only classified correctly 29.80\% of the time, and most often misclassified as standing 43.20\% of the time. Through searching we find using a simpler network with fewer weights generalizes better with more equal class probabilities, but does not achieve a higher total accuracy. \autoref{fig:model_search_epochs} illustrates the impact of using a deeper network structure with an increasing number of convolutional filters. It is evident that the network begins to overfit the training data after the first 10 epochs as the validation set begins to plateau. To better understand these results, we investigate what the model has learned by exploring explainable methods and discuss this in \autoref{sec:model_interpretation}.

\subsection{Transfer Learning}
A close relationship exists between the complexity of a model and the data distribution. With a limited number of data samples, we find that a simpler model with fewer parameters is desired, but lacks in generalizing well to unseen images. Therefore, we justify the use of transfer learning as a way to fine-tune larger models trained on ImageNet and improve generalization for our dataset. For MobileNetV2 and family of EfficentNet models, we observe higher classification accuracies for all classes. Both architectures take as input 224x224 RGB images and output the probability of each class. After removing the output layer of the pre-trained model, we append a global average pooling layer to flatten the output, and then using a dense layer with the ReLU activation, followed by dropout with a probability of 0.35, batch normalization with default parameters, and then the final output layer with the softmax activation. The remaining hyperparameters are identical to ConvNet with the exception of using a learning rate of 0.00001 and only training for 20 epochs. 

While searching for the highest accuracy transfer learning model we encountered overfitting conditions as training epochs were increased. By testing various regularization techniques, such as dropout and batch normalization, we find slight improvements to reduce overfitting. However, having too high a dropout probability results in lower overall accuracy and the lack of ability to converge to a local minima. Furthermore, we believe that with more labeled images the transfer learning models could be improved further. We applied several augmentations to the dataset in an effort to increase the available sample size to help with overfitting. Still, the accuracy of the transfer learning model can be improved upon if an abundance of additional labeled data was included beyond the Stanford dog breed dataset. 

The results of using transfer learning yield a classification accuracy on the test set of 87.52\% and 92.06\% for MobileNetV2 and EfficentNetB4, respectively. Training takes 13 minutes for MobileNetV2 and 52 minutes for EfficientNetB4, which is significantly greater as a result of increased complexity in network topology. While EfficientNetB4 provides the highest overall accuracy, it is not the ideal candidate for our application, as discussed in  \autoref{sec:results}.  The additional four EfficientNetB0-3 models have a range of values seen between these boundaries. 

\section{Model Interpretation}
\label{sec:model_interpretation}
We show the use of deep learning as a successful method to discern the behavior of dogs through minimally processed images. However, the high dimensional and non-linear nature of neural networks make the results difficult to interpret. As a result we assess two methods; Gradient-weighted Class Activation Mapping (GradCAM) \cite{Selvaraju2019GradCAMVE} and Integrated Gradients \cite{Sundararajan2017AxiomaticAF}, to study how the input features attribute specific predictions and make the models more transparent.  

\begin{figure}[t]
\includegraphics[width=8.7cm]{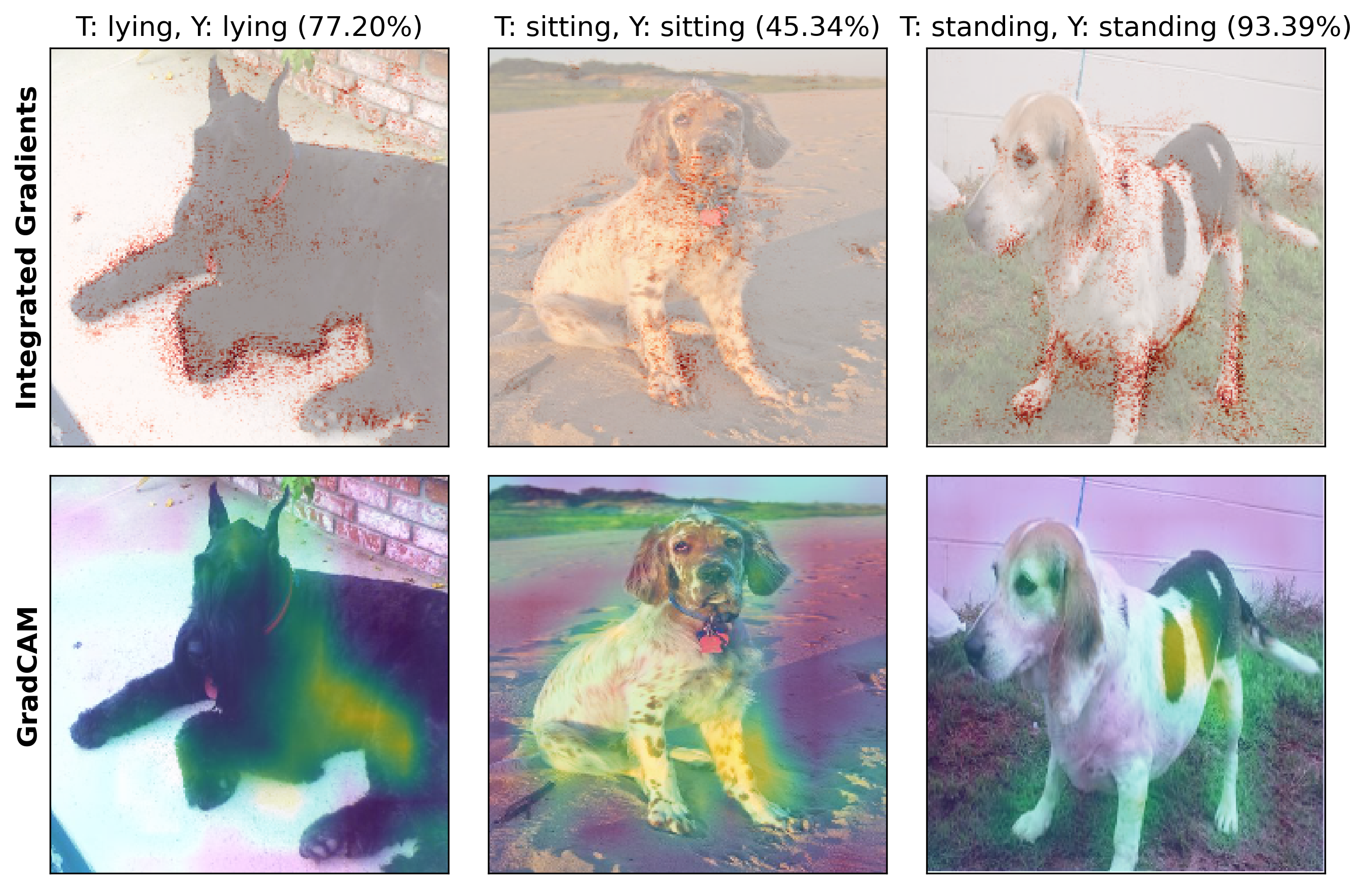}
\caption{Two interpretability methods for the three classes. `T' represents the ground truth label, `Y' is the model prediction followed by the softmax probability for that class in parentheses.}
\label{fig:interpretability_methods}
\end{figure}

From each of the three classes we extract a single image that the network correctly predicts, and evaluate each with the two methods using ConvNet. Understanding the hand-crafted model provides insight to not only what areas of the image are important for prediction, but also why it may struggle identifying the sitting behavior. \autoref{fig:interpretability_methods} displays the results of both methods overlaying the original input image. The integrated gradients are obtained by accumulating all the gradients along a straight line path from a baseline zero-intensity image to the input sample. It can be seen that the visualizations of integrated gradients pick up distinctive features for the lying and standing images well. That is, the input pixels surrounding the lower body and legs of the dog are most important when inferring a prediction. However, the image of the dog sitting does not have integrated values with high intensity.   

Integrated gradients highlight fine-grained details at the pixel-level. Alternatively, GradCAM localizes relevant regions in an input image through the activation maps of successive convolutional layers using the desired class signal and back propagating gradients to compute a localization heat map. The bottom row of \autoref{fig:interpretability_methods} shows which areas of the image the network looks at to make a prediction. We are confident that our model is learning as expected by capturing the position and posture of the dog to infer the scene. Additionally, we observe that images labeled as lying down tend to have horizontal regions with high importance. A similar result is seen with standing dogs having multiple vertical regions overlaying on the dog’s legs. Consequently, sitting dogs have a combination of the two features, with vertical and horizontal regions of importance. 

\section{Experimental Evaluation}
\label{sec:results}
We analyze a number of desired traits in order to determine the best models beyond the standard network performance metrics. Specifically, we evaluate the model size and inference time of trained neural networks. We determine that accuracy of the network is most important to our application, inference speed secondly, and model size lastly. We find the model size to be least important because of the Jetson Nano's large VRAM, relative to other embedded devices without dedicated hardware accelerators. With the Nano’s capabilities we can afford an amount of flexibility with model compression.  

In each experiment we explore two model optimizations; TensorFlow Lite and TensorRT in 16-bit and 32-bit compression factors. We find that TensorFlow Lite achieves better compression and contributes to overall smaller models, while TensorRT achieves better performance and contributes to higher frame rates. The spread of model sizes ranges from 2.32 MB to 110 MB, while the frame rate performance ranges from less than 1 to 39 FPS.  

\begin{table}[t]
\caption{Inference Speed Comparison [FPS]}
\label{tab:inference_speed}
\begin{tabularx}{\columnwidth}{@{}l*{2}{C}c@{}}
\toprule
networks       & baseline & tflite* & tensorrt*  \\ 
\midrule
ConvNet        & 2.134    & 7.383   & 39.014  \\
MobileNetV2    & 1.339    & 9.381   & 33.498  \\ 
EfficientNetB0 & 1.617    & 2.895   & 36.281  \\ 
EfficientNetB1 & 0.923    & 1.995   & 24.863  \\ 
EfficientNetB2 & 0.552    & 1.843   & -----   \\ 
EfficientNetB3 & 0.042    & 1.353   & -----   \\ 
EfficientNetB4 & -----    & 0.963   & -----   \\ 
\bottomrule
\end{tabularx}
* models with the highest inference speed's are displayed.
\end{table}

\subsection{Baseline Inference Speed}
Inference times on the Jetson Nano are determined primarily by the optimization method that is chosen. \autoref{tab:inference_speed} displays the results for each model when using no optimizations and comparing it to the best case TensorFlow quantization and TensorRT optimizations methods. All results shown with values are considered valid methods, whereas those with the missing values were not able to be run successfully on the Nano due to the device running out of memory and crashing. The key takeaway from the table shows that when large models are unable to run they benefit from quantization methods. What's more, the use of TensorRT will increase the inference times by at most 27x. The following subsections discuss the specific performance metrics and comparisons for each optimization method in greater details.

\begin{figure}[t]
\includegraphics[width=8.8cm]{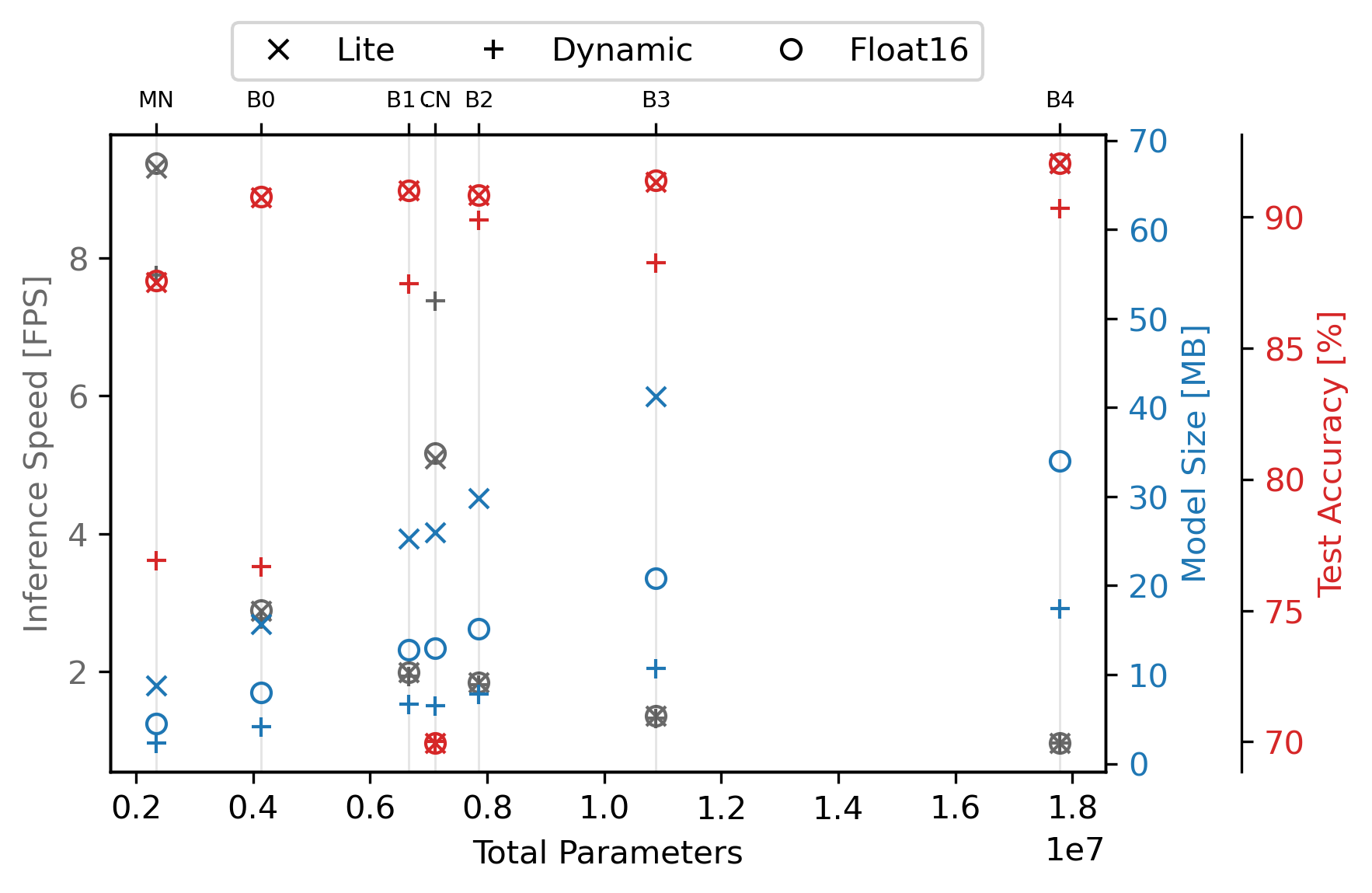}
\caption{Model details and performance results from three post training quantization methods (i.e. standard, dynamic, and 16-bit floats) on ConvNet (CN), MobileNetV2 (MN), and EfficientNet (B0-4). The number of parameters, inference speed on the Jetson Nano, memory footprint of the model, and test accuracy are considered.}
\label{fig:tflite_model_eval}
\end{figure}

\subsection{TensorFlow Lite Results}
\autoref{fig:tflite_model_eval} provides a comprehensive overview on the influence of each quantization method from TensorFlow for the seven best models found during the design search. The inference speed shown is captured on the Jetson Nano by running inference on 64 batches of one image and dividing the total time by the number of batches. Note that a randomly generated image is cached for the test and the inference speed is not indicative of how long it takes for the camera to capture an image. 

The first model we explore is transfer learning using MobileNetV2. We find the benchmark performance for the 16-bit TensorFlow Lite variation performs adequately at 9.38 FPS but has a memory footprint of 4.48 MB when used with 16-bit floating point quantization. When using dynamic quanitzation the model size can be reduced to 2.38 MB, but suffers from a significant decrease in accuracy of nearly 10 percentage points. A similar behavior can be observed for all the other models when using the dynamic method. This offers a significant tradeoff between inference time and model size, but our application need not exceed 10 FPS to operate. Overall, MobileNetV2 proves to be the best middle-ground model for our application thus far.

The second most efficient model with respect to inference speed, is the ConvNet model. However, as \autoref{fig:tflite_model_eval} shows, the model suffers from having the lowest test accuracy deeming it insufficient for our application. As a result we find the remaining transfer learning models to be more important to compare.

Lastly, we compare the family of EfficientNet models which collectively have the highest accuracy but worst inference speed. The decrease in speed is directly correlated to the number parameters in the model and follow decrease logarithmically regardless of the quantization method used. Alternatively, choosing a quantization method is more complex. The change in accuracy between B0 and B3 is marginal with a change of accuracy within 2 percentage point difference in best case scenarios. The benefits to using either method will depend on whether smaller memory footprint or faster inference times is more important. If accuracy is preferred over inference speed, then the EfficientNetB0 variant when used with 16-bit floating point weights is the best option. Specifically, the model achieves the 90.79\% test accuracy, at 2.89 FPS, and only has memory footprint of 7.98 MB.

\begin{figure}[t]
\includegraphics[width=8.8cm]{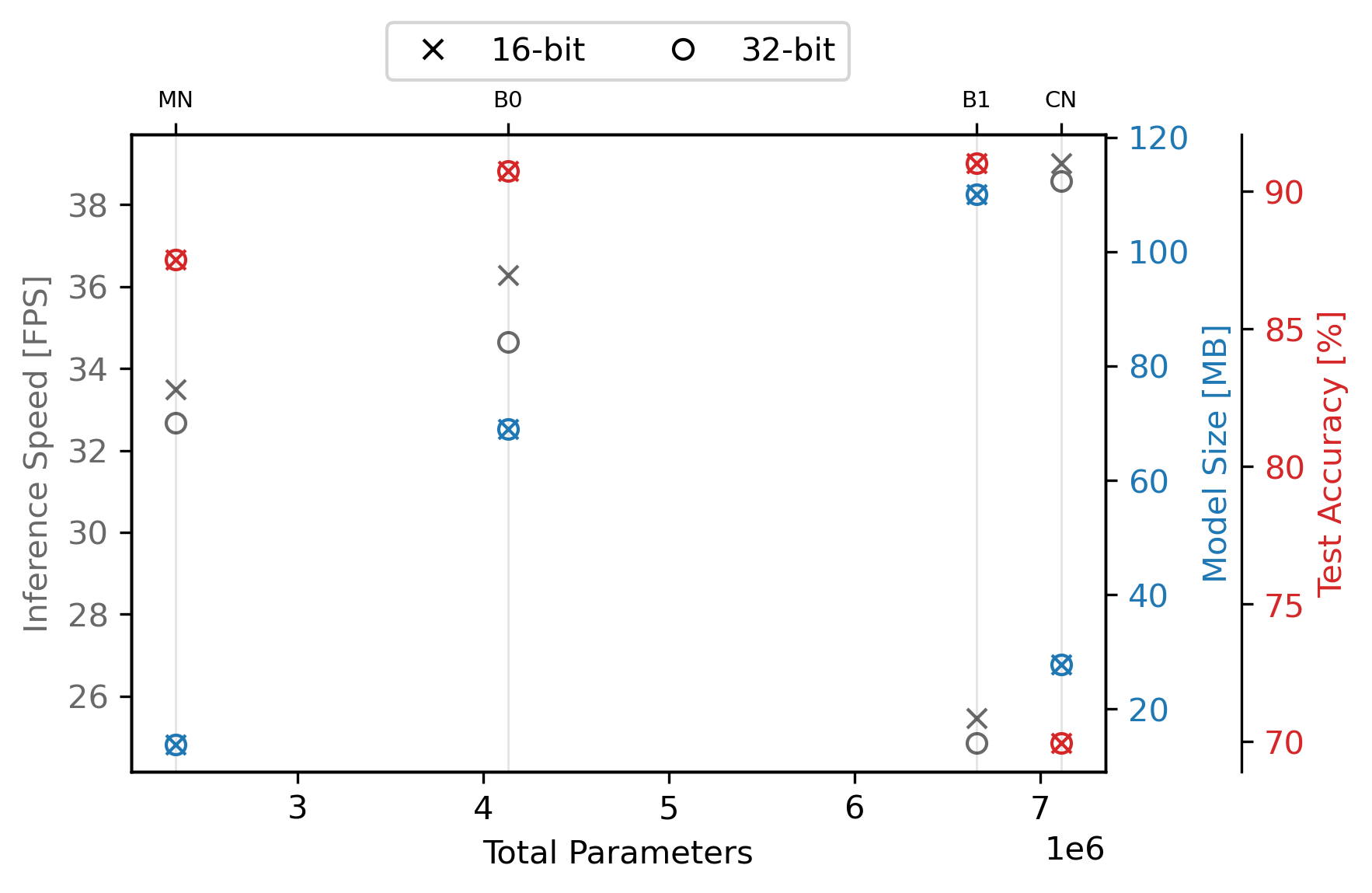}
\caption{Model details and performance results from 16-bit and 32-bit TensorRT optimizations on ConvNet (CN), MobileNetV2 (MN), and EfficientNet (B0,1). The number of parameters, inference speed on the Jetson Nano, memory footprint of the model, and test accuracy are considered.}
\label{fig:trt_model_eval}
\end{figure}

\subsection{TensorRT Results}
A key benefit to using TensorRT is the increase in inference speed performance. The results from utilizing this optimizer with the Jetson Nano show a dramatic increase of up to 27x speed up over it's unoptimized counter-parts. Relative to the baseline figures, the optimized models do not result in a memory footprint reduction or change in accuracy. For our research purposes we find that an increase in inference speed was less important than accuracy, and model memory footprint was crucial for being able to run on the Jetson Nano. For other applications that require inference speed we would recommend using TensorRT. However, TensorRT is not perfect for all scenarios. For example, the TRT engines were unable to build EfficientNet B2-4 models and thus were unable to run due to the memory constraints of the Jetson Nano's GPU.

As observed in the experiments, \autoref{fig:trt_model_eval} illustrates the performance between ConvNet, MobileNetV2, and EfficentNet TensorRT optimization results. Our hand designed ConvNet model achieves the highest inference speed at 39.01 FPS with the 16-bit optimization, with the 32-bit version coming in lower at 38.57 FPS. However, performance is significantly worse than other models in test accuracy at 69.95\%, with a memory footprint of 27.8 MB for both 16 and 32-bit models, respectively. ConvNet is the best performing model in terms of inferences per second. Although, this comes at the expense of accuracy as the model does not learn to classify ``sit'' well enough to be considered the best overall model. Furthermore, ConvNet has 3.5x more total parameters than the smallest, which is MobileNetV2. 

MobileNetV2 optimized with TensorRT for both 16 and 32-bit achieves an inference speed of 32.67 and 33.49 FPS, respectively. The accuracy comes in at 87.52\% and has a model size of around 13.8 MB for both 16 and 32-bit. Although MobileNetV2 is the smallest model, the accuracy suffers a bit compared to the EfficientNet models. 

The EfficientNetB0 16-bit version performs at 36.28 FPS while the 32 bit version performs at slightly lower at over 34.65 FPS. Both 16 and 32-bit versions have a 90.75\% test accuracy and a memory footprint of 69 MB. This model contains the largest spread for inference speed between 16 and 32-bit precision. For EfficientNetB1, we observe a large inference speed drop at around 25.46 FPS for 16-bit and 24.86 FPS for the 32-bit. Test accuracy climbs to 91.03\%, and carries the largest model memory footprint at 110 MB. The B1 model has nearly 1.9x as many parameters as compared to B0, but shows a major drop in inference speed. However, with consideration towards model size and inference speed, we determine EfficientNetB0 as the best model to use when using TensorRT.

\section{Discussions}
Some difficulties we observe with rewarding ``sit'' or ``lie'' behaviors include rapid change between inference results. For example, when using a model with an inference time of 30 FPS we do not want to produce a treat based on every frame we get a result on. Such implementation results in a rapid delivery of many treats. Additionally, we want to protect against the observation of false positives as we do not want to reward a behavior that are demonstrated by the dog for only a split second or in between actions. We address these cases with a variety of protections which allow for smooth operation. When an inference is made on an image, the device responds in one of two methods; dropping treats or no action. A FIFO circular buffer is populated with the classification result of each inference. We make a determination to deliver a treat based on the number of inferences in the queue which share a majority result. This approach allows the machine to determine if the dog is displaying a particular behavior for a given length of time.

In comparison of models we encounter a trade off between memory footprint, inference speed, and classification accuracy. The ideal condition is to produce a model which excels in all three categories. We have found that models which perform with the highest classification accuracy may have a larger memory footprint. Models that have fewer parameters will show an increase in inference speed. The benefits of each category is dependent on the specifications of the final application. For the current study, the criteria for which model is best is a balance between accuracy and inference speed. Regardless of how efficient a model runs, we are ultimately constrained to a frame rate of 30 FPS as captured by the camera. However, we find our application operates smoothly with a frame rate at around 10 FPS. Any frame rate above this provides no significant benefit for the use of the application to deliver treats.

\section{Conclusion}
Using the techniques discussed in this paper we demonstrate how to maximize the efficacy of our models and showcase improvements in accuracy and performance for various neural network architectures. Additionally, we identify key trade-offs between using various quantization and optimization methods through a range of extensive experiments. In particular, we present two primary use cases when designing a neural network for reinforcing positive behavior in dogs on an embedded device. The first emphasizes inference speed, while the other optimizes for reducing memory size. We consider the following enhancements to our methodology as future work: including additional classes of dog actions and implementing Object Detection to improve the recognition of subtle dog behaviors (e.g., the positioning of ears and tails). This may provide a deeper knowledge of the dog's emotional state in addition to it's body posture.

\bibliographystyle{IEEEtran}
\bibliography{refs.bib}

\end{document}